\documentclass[10pt,twocolumn,letterpaper]{article}

\usepackage{cvpr}              

\usepackage{graphicx}
\usepackage{amsmath}
\usepackage{amssymb}
\usepackage{booktabs}
\usepackage[ruled,vlined]{algorithm2e}
\newtheorem{definition}{Definition}{}
\usepackage{xcolor}
\usepackage{multirow}
\usepackage[pagebackref,breaklinks,colorlinks]{hyperref}
\usepackage[accsupp]{axessibility}

\usepackage[capitalize]{cleveref}
\crefname{section}{Sec.}{Secs.}
\Crefname{section}{Section}{Sections}
\Crefname{table}{Table}{Tables}
\crefname{table}{Tab.}{Tabs.}

\def\cvprPaperID{7704} 
\def\confName{CVPR}
\def\confYear{2022}

\begin{document}

\title{Arch-Graph: Acyclic Architecture Relation Predictor for Task-Transferable Neural Architecture Search}
\author{Minbin Huang$^{1}$ \quad Zhijian Huang$^{1}$ \quad Changlin Li$^{3}$ \quad Xin Chen$^{4}$ \quad Hang Xu$^{2}$\\ Zhenguo Li$^{2}$ \quad Xiaodan Liang$^{1}$\thanks{Corresponding author.}\\
{\normalsize $^{1}$Shenzhen Campus of Sun Yat-sen University \quad $^{2}$Huawei Noah's Ark Lab}\\
{\normalsize $^{3}$ReLER, AAII, UTS \quad $^{4}$The University of Hong Kong}\\
{\tt\small \{huangmb5,huangzhj56\}@mail2.sysu.edu.cn, 	changlinli.ai@gmail.com, cyn0531@connect.hku.hk,}\\
{\tt\small  chromexbjxh@gmail.com, 	li.zhenguo@huawei.com, xdliang328@gmail.com}
}
\maketitle
\begin{abstract}
\vspace{-2mm}
Neural Architecture Search (NAS) aims to find efficient models for multiple tasks. Beyond seeking solutions for a single task, there are surging interests in transferring network design knowledge across multiple tasks. In this line of research, effectively modeling task correlations is vital yet highly neglected.
Therefore, we propose \textbf{Arch-Graph}, a transferable NAS method that predicts task-specific optimal architectures with respect to given task embeddings. It leverages correlations across multiple tasks by using their embeddings as a part of the predictor's input for fast adaptation.
We also formulate NAS as an architecture relation graph prediction problem, with the relational graph constructed by treating candidate architectures as nodes and their pairwise relations as edges. To enforce some basic properties such as acyclicity in the relational graph, we add additional constraints to the optimization process, converting NAS into the problem of finding a Maximal Weighted Acyclic Subgraph (MWAS). Our algorithm then strives to eliminate cycles and only establish edges in the graph if the rank results can be trusted.
Through MWAS, Arch-Graph can effectively rank candidate models for each task with only a small budget to finetune the predictor.
With extensive experiments on TransNAS-Bench-101,
we show Arch-Graph's transferability and high sample efficiency across numerous tasks, beating many NAS methods designed for both single-task and multi-task search.
It is able to find top 0.16\% and 0.29\% architectures on average on two search spaces under the budget of only 50 models.\footnote{%
Code: {\scriptsize\url{https://github.com/Centaurus982034/Arch-Graph}}}
\end{abstract}
\begin{figure}
    \centering
    \includegraphics[width=\columnwidth]{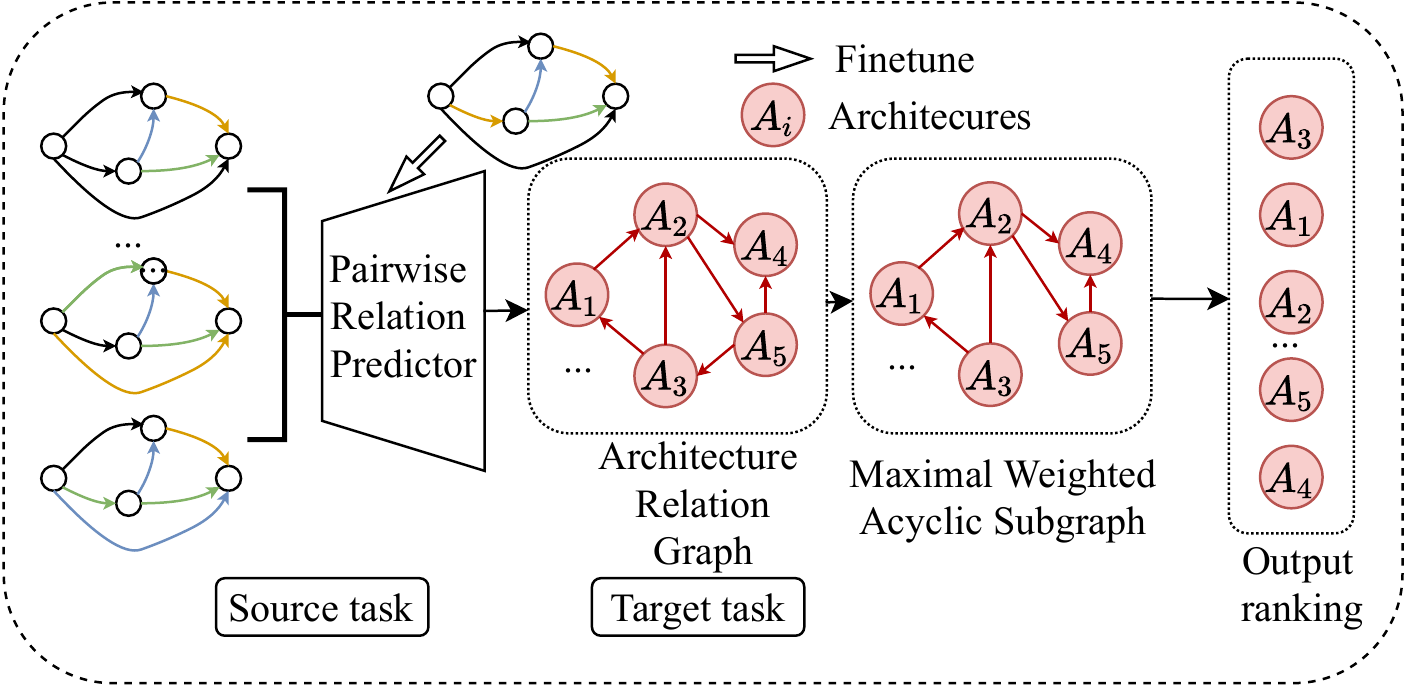}\vspace{-3mm}
    \caption{Overview of our Arch-Graph that trains a pairwise relation predictor on a source task and transfers to target task by finetuning. It constructs an architecture relation graph based on the pairwise relation predictor. After selecting the MWAS of the architecture relation graph, Arch-Graph can give a proper ranking of different candidate architectures.}
    \label{fig:overview1l}\vspace{-6mm}
\end{figure}
\vspace{-5pt}
\section{Introduction}
\label{sec:intro}







Neural Architecture Search (NAS) methods\cite{zoph2018learning,DBLP:conf/iclr/BakerGNR17} have the potential to democratize deep learning and reduce costly human labor in designing neural networks. By automatically searching for optimal architectures, many NAS methods have discovered models exceeding human-designed ones on various tasks. However, many NAS solutions are computationally expensive as they require training over numerous candidate architectures. Under cases where networks for multiple tasks are needed, searching for an architecture for each task requires repeatedly running NAS methods from scratch to find the top performing network, throwing away potentially valuable knowledge accumulated over the course of searching.
 There are many recent attempts\cite{DBLP:conf/nips/WongHLG18, DBLP:conf/iclr/LeeHH21} investigating transferable NAS problems over different tasks by mining task correlations. For instance, \cite{DBLP:conf/iclr/LeeHH21} proposes to use meta-learning to generate architectures for a given new task. However, it makes a strong assumption that information on top-performing architectures for each pretrain task is always available, which can limit its use case. \cite{DBLP:conf/nips/WongHLG18} proposed to use task embeddings to inform an RNN controller of the task information and framed NAS as a reinforcement learning (RL) problem, which inherits the sample inefficiency problem from RL. Weight-sharing techniques\cite{DBLP:conf/icml/PhamGZLD18, DBLP:conf/iclr/LiuSY19,DBLP:conf/cvpr/LiPYWLLC20,DBLP:conf/iccv/LiTWPWLC21} are recently more popular among researchers due to their efficiency in cost reduction, typically by training a supernet and then inheriting weights from it. However, due to their restrictions in supernet design, weight-sharing methods are usually constrained in the choice of network search space.

Predictor-based NAS methods\cite{DBLP:conf/eccv/WenLCLBK20, DBLP:conf/eccv/NingZZWY20, DBLP:conf/cvpr/Xu00TJX021, DBLP:conf/nips/ShiPXLKZ20, wu2021weak} alleviate these concerns by sampling architecture-performance pairs and fitting a proxy accuracy predictor to reduce computation costs. However, training a large number of architectures for fitting a good predictor can also be computationally challenging. Besides, this approach is ultimately converting NAS into a regression problem, which can be hard to solve since the model space is usually highly non-convex, making accurately identifying top performers extremely difficult. In this paper, we instead argue that approaching NAS as a \textit{ranking} problem can bring along many extra benefits compared to other methods, largely due to its added constraints that provide extra learning signals. 

This key observation motivated us to develop a predictor that captures pairwise relations among architectures and formulate NAS as a graph ordering problem. Our method, Arch-Graph, treats architectures as nodes and order information as directed edges, such that an edge pointing from $\mathit{arch}_a$ to $\mathit{arch}_b$ represents the superiority of $\mathit{arch}_a$ in its performance when compared to $\mathit{arch}_b$. We propose to use a pairwise relation predictor to construct this graph. This predictor is optimized with objective of finding the correct \textit{pairwise} order of nodes in the graph, which greatly improves data efficiency and prediction accuracy comparing to previous pointwise predictor that directly predict architecture performance. 

To allow transfering among different tasks without re-training the predictor, another key ingredient \textit{task embedding} that represents a task during the predictor training process stablizes the knowledge transfer between tasks. Previous works on task embedding mostly focus on classification tasks~\cite{DBLP:conf/iccv/AchilleLTRMFSP19}, whereas our proposed task embedding method is more general and can be applied to many other vision domains such as autoencoding and semantic segmentation.


After constructing the relation graph through the pairwise predictor, the architecture selection can then be formulated as a topological ordering problem on this graph. Under this setting, it is vital to enforce that the graph follows basic properties of a partial order, such as acyclicity, which prohibits circular ordering (A $>$ B $>$ C while C $>$ A). Therefore, a central component of our work is the definition of a Maximal Weighted Acyclic Subgraph (MWAS) problem with Trust Score to make sure the constructed graph follows the irreflexive, transitive, and anti-symmetric properties of a partial order. We propose an approximation solution to it by iteratively applying the \emph{max-MAS} algorithm. 



Our experiments on TransNAS-Bench-101 proves the effectiveness of Arch-Graph, identifying architectures with average rank 5.24 (top 0.16\%) and 12.2 (top 0.29\%) on macro and micro search space respectively, with only randomly sampling 50 architectures, saving at least 37.5\% of samples in other methods to achieve comparable results.



To conclude, the contributions of our work can be summarized as follows:
\vspace{-2mm}
\begin{itemize}
\item We propose Arch-Graph, a task transferable NAS method by formulating NAS from a novel perspective: A graph ordering problem, and
solve this problem by training a \textit{pairwise} relation predictor, which is more data efficient, saving at least 37.5\% training samples.
\vspace{-2mm}
\item We generalize \textit{task embeddings} to any kind of tasks, and further enables task-transferable NAS by predicting architecture relation on any given task embeddings. 

\vspace{-2mm}
\item To remove incorrect edges in the relation graph constructed by the predictor, we define the Maximal Weighted Acyclic Subgraph problem and propose an approximation algorithm to solve it. 
\vspace{-2mm}
\item Extensive experiments demonstrate that Arch-Graph can beat many existing transferable NAS methods by a large margin, finding top 0.16\% and 0.29\% architectures on two search spaces. 
\end{itemize}
\vspace{-2mm}

\begin{figure*}
    \centering
    \includegraphics[width=2\columnwidth]{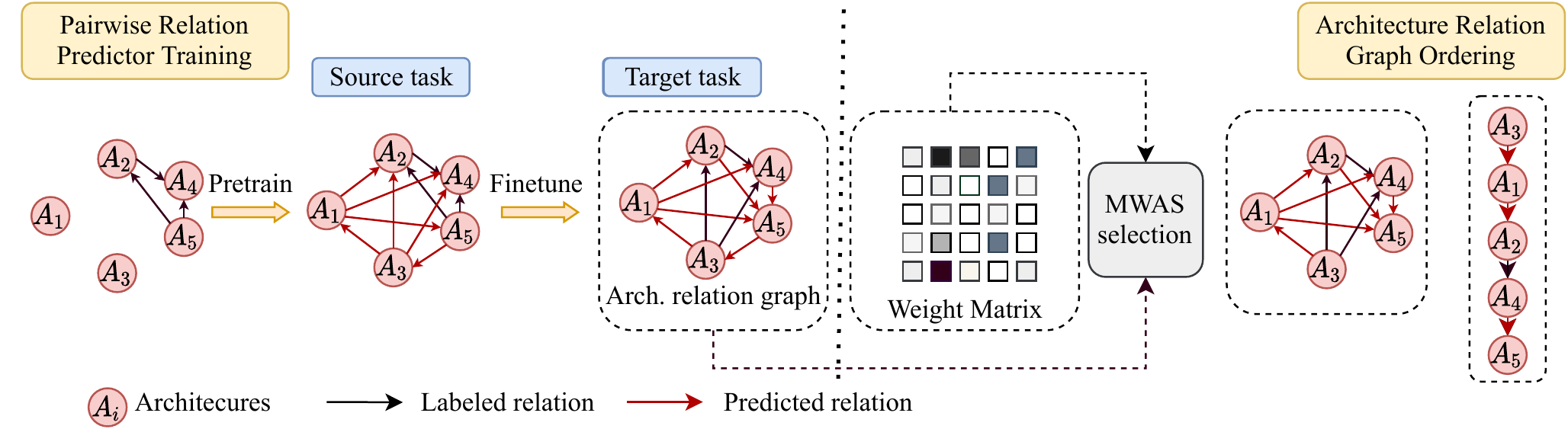}
    \vspace{-3mm}\caption{Framework of our proposed Arch-Graph. In the Pairwise Relation Predictor Training stage, given a source task and architectures of interest, we sample a small budget of architectures to fit the predictor then finetune it on a target task. Next, in the Architecture Relation Graph Ordering stage, we construct a relation graph according to the prediction, treating each architecture as a node and directed edge as ranking information. To get a proper ordering from the relation graph, we assign weights related to the confidence to the edges and select the Maximum Weighted Acyclic Subgraph and get a Directed Acyclic Graph (DAG) from the relation graph. Finally, we evaluate the top architectures given by the topological sorting of this DAG. 
    }
    \label{fig:overview}
    \vspace{-3mm}
\end{figure*}
\section{Related Work}
\label{sec:related work}

\noindent\textbf{Predictor-based NAS.} NAS has achieved many breakthroughs in the past few years. Its early works utilized reinforcement learning \cite{DBLP:conf/cvpr/TanCPVSHL19, DBLP:conf/cvpr/ZhongYWSL18,  DBLP:conf/iclr/ZophL17, DBLP:conf/cvpr/ZophVSL18} and evolutionary algorithms\cite{DBLP:conf/aaai/RealAHL19, DBLP:conf/icml/RealMSSSTLK17, DBLP:conf/icml/SuganumaOO18, DBLP:conf/iccv/XieY17, DBLP:journals/corr/abs-1810-03522} and found many top-performing architectures at a high computational cost. Later works then strive to reduce the search cost while improving performance. Among numerous directions, predictor-based NAS methods are most relevant to our work. They try to predict the performance of a given neural architecture both accurately and efficiently. These methods usually involve two steps: 1) Sampling pairs of architectures and their accuracies, and 2) learning the accuracy predictor. 
The objective of fitting the predictor can be regarded as a regression\cite{DBLP:conf/eccv/WenLCLBK20} or ranking \cite{DBLP:conf/eccv/NingZZWY20, DBLP:conf/cvpr/Xu00TJX021} problem, and there is a wide range of choices for predictors \cite{DBLP:journals/corr/abs-2003-12857, DBLP:conf/nips/LuoTQCL18, DBLP:conf/nips/DudziakCALKL20, DBLP:journals/corr/abs-2007-04785}. 
Shi \etal \cite{DBLP:conf/nips/ShiPXLKZ20} adopted a bayesian sigmoid regression as the surrogate model for Bayesian Optimization (BO) to select candidates. As applying BO on the whole search space is difficult, weakNAS\cite{wu2021weak} replaced one strong predictor with a set of weaker predictors to get oversimplified BO. Different from these previous works, we propose pairwise relation predictor and formulate NAS problems as a graph ordering problem where the graph is given by the predictor.

\noindent\textbf{Transferable NAS.} Transfer learning for NAS mainly focuses on transferring between tasks using the same search space and between search space on a specified task. There are some recently proposed cross-task NAS benchmarks\cite{DBLP:conf/cvpr/DuanCXCLZL21, DBLP:journals/corr/abs-2110-05668} for improving the transferability and generalizability of NAS algorithms. Though relatively neglected when compared to single-task NAS, there are still some outstanding algorithms. CAS \cite{DBLP:conf/acl/PasunuruB19} applies continuous learning on multi-task architecture search based on a weight sharing strategy, trying to find a single cell structure that can generalize well to unseen tasks.
Catch\cite{DBLP:conf/eccv/ChenDCXCLZL20} combined meta-learning with RL to swiftly adapt to new tasks.
Different from \cite{DBLP:conf/nips/LuoTQCL18} for single task, Lee \etal \cite{DBLP:conf/iclr/LeeHH21} proposed to generate graphs from datasets in a meta-learning style to make the methods generalize well across multiple datasets. However, it requires top-performing architectures during training to learn characteristics of good models, which can incur high computational costs. Contrary to this, our method achieves remarkable results only by random sampling. 


\section{Arch-Graph}
Transferable NAS methods aim to reuse the architecture selection knowledge from source tasks and find top-performing architectures on a target task.
Consistent with this setting, the Arch-Graph algorithm consists of two parts: pairwise relation predictor training and architecture relation graph ordering, as illustrated in \cref{fig:overview}. We train the pairwise relation predictor (\cref{sec:predictor}) on the source task using sampled architecture pairs and task embeddings (\cref{sec:task embedding}), then finetune it on the target task. After constructing architecture relation graphs using the finetuned predictor, we rank the architectures by finding a Maximum Weighted Acyclic Subgraph (MWAS) (\cref{arch-graph prediction}).

\label{sec:methods}

\subsection{Pairwise Relation Predictor}\label{sec:predictor}

For a predictor-based NAS algorithm, the predicted ranking of models might matter more than absolute numbers of model performance prediction, since we only care about the top-ranking ones. 
Many predictor-based NAS methods focus on directly predicting the accuracy of models\cite{DBLP:conf/eccv/WenLCLBK20} or the ranking of all models of interest through a ranking loss\cite{DBLP:conf/cvpr/Xu00TJX021,DBLP:conf/eccv/NingZZWY20}. However, since model spaces are usually highly non-linear, these predictors typically cannot be trained to have high accuracy. Moreover, these methods are not data-efficient since they need lots of samples to fit the predictor on a complicated model space.

We propose to study NAS from a new perspective, which is to formulate it as an architecture relation graph ordering problem.
Our key observation is that while ranking all models can be problematic, it is much easier to make comparisons just between two models. Besides, as previous works ~\cite{pmlr-v139-feng21d,NEURIPS2020_768e7802} illustrated, when challenged with limited available data, learning pairwise relations can yield a higher classifier performance than many common regression methods. This is because we can "augment" the data by constructing $n^2-n$ pairs of relations when we only have $n$ labeled samples. This is extremely helpful in settings where obtaining labels is computationally expensive, such as NAS. This inspired us to use a well-trained pairwise relation predictor to get a ranking of models in a search space. It is thus crucial to properly define \emph{relation} in our settings, where the most relevant concept is \emph{partial order}.
\vspace{-2mm}
\begin{definition}
\label{def_partial}
(partial order)
A (strong) partial order on a set $P$ is a relation $\prec$ that is both \textbf{irreflexive}, \textbf{transitive} and \textbf{anti-symmetric}, that is, for $\forall a,b,c \in P$:\\
1. irreflexive: not $a \prec a$. \\
2. transitive: if $a \prec b$ and $b \prec c$ then $a \prec c$.\\
3. anti-symmetric: if $a \prec b$ then not $b \prec a$.
\end{definition}
\vspace{-5mm}
\begin{definition}
\label{def_total}
(total order)
a total order is a partial order on a set $P$ so that for $\forall a,b\in P$, either $a \prec b$ or $b \prec a$.
\end{definition}
\vspace{-2mm}
If a well-trained predictor defines a partial order, the problem of ranking models is then reduced to extending a partial order (\cref{def_partial}) to a total order (\cref{def_total}), which has been extensively studied in the existing literature. Therefore, our predictors are trained to define a partial order on the model space.

\begin{figure}
    \centering
    \includegraphics[width=\columnwidth]{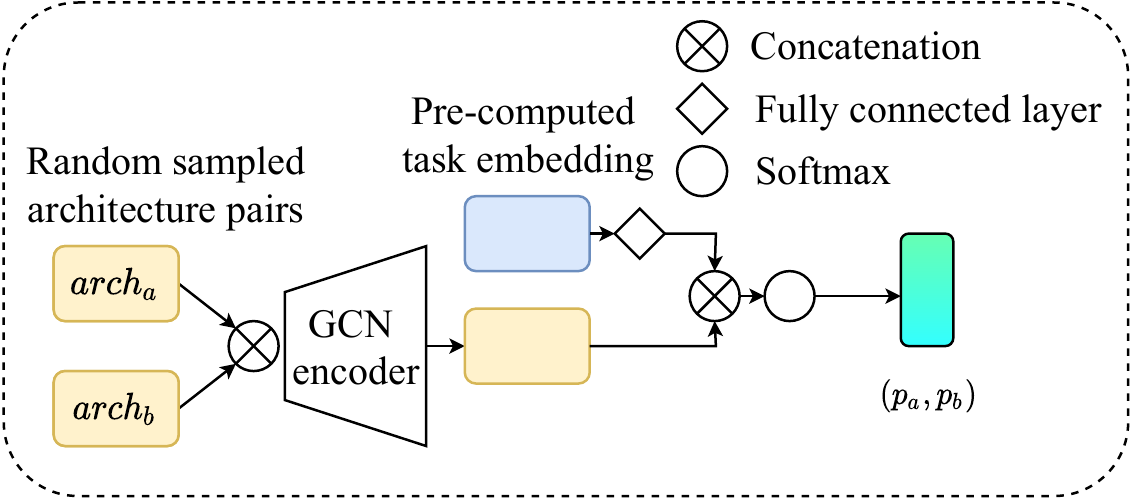}\vspace{-3mm}
    \caption{Detailed structure of our proposed pairwise relation predictor. The predictor takes an architecture pair $(\mathit{arch}_a,\mathit{arch}_b)$ and a task embedding as input and produce a probability vector 
    $p_a,p_b$, where$p_a>p_b$ indicates that $\mathit{arch}_a$ is better than $\mathit{arch}_b$.}
    \vspace{-3.5mm}
    \label{fig:predictor}
\end{figure}

Given a source task $\tau_s$, we first randomly pick $m$ models from $\tau_s$ and fully evaluate them to get their performance on the test dataset. In this way, we obtain $m^2-m$ samples by forming pairwise relations. Details of the pairwise relation predictor is illustrated in \cref{fig:predictor}.
The $(\mathit{arch}_a,\mathit{arch}_b)$ are randomly sampled architectures that are first concatenated as the input of a Graph Convolutional Network (GCN)\cite{DBLP:conf/iclr/KipfW17}predictor. The GCN predictor then generates two embeddings to represent these two architectures. Next, these embeddings are concatenated with a task embedding, which is generated by applying a fully connected layer to the feature extractor described in \cref{sec:task embedding}. Together, they are fed into a softmax function to construct a simple probability distribution $p=(p_a,p_b) \in \mathbb{R}^2$ with $p_a > p_b$ indicating $\mathit{arch}_a$ is better than $\mathit{arch}_b$. The produced probability distribution is then compared with the ground truth label\{$[0,1]^T,[1,0]^T$\}. The objective is to minimize the Binary Cross Entropy (BCE) Loss. Specifically, we include both $(\mathit{arch}_a,\mathit{arch}_b)$ pairs and $(\mathit{arch}_b,\mathit{arch}_a)$ pairs to encourage anti-symmetry. If neither $a \to b$ nor $b \to a $ exists, we simply mark them as incomparable, which is allowed in a partial order.

After training the pairwise relation predictor on a source task, we conduct transfer learning by finetuning the predictor on a set of $t$ target tasks $\{\tau_1,\tau_2,...,\tau_t\}$ with a small budget of $b$ architectures chosen from each target task. More specifically, $b_f$ architectures for finetuning the predictor and $b_v$ architectures for pairwise relation validation. We pick the predictor with the highest validation accuracy as the final result. Then, the architecture relation graph ordering is performed on $\tau_i$ on top of the finetuned predictor.

\subsection{Task Embedding}\label{sec:task embedding}
When transferring architecture knowledge across tasks, it is important to inform NAS methods of the target task's intrinsic characteristics and adjust the architecture selection strategy accordingly. We therefore follow ~\cite{DBLP:conf/iccv/AchilleLTRMFSP19}, which only generates task embeddings for classification tasks, extend it to generate embeddings for other tasks.

A task's nature can be quantified by the neural network's weights when trained on this task.
When a pre-trained model is finetuned on a task $\tau_i$, it is actually 
adding some perturbation $w^{\prime}=w + \delta w$ to a network's weights and we can measure the average KL divergence between the original output distribution $p_{w}(y|x)$ and the perturbed one  $p_{w'}(y|x)$. It can be measured by
\begin{equation}
    \mathbb{E}_{x \sim \hat{p}}KL(p_{w'}(y|x)||p_w(y|x))=\delta w F \delta w
\end{equation}
where F is the Fisher Information Matrix (FIM):
\begin{equation}
    F=\mathbb{E}_{x, y \sim \hat{p}(x) p_{w}(y|x)}\left[\nabla_{w} \log p_{w}(y|x) \nabla_{w} \log p_{w}(y| x)^{T}\right]
\end{equation}
The FIM then indicates the set of feature maps which are more informative for solving the current task.
We use an ImageNet pre-trained ResNet-50 as the encoder, then train an encoder-decoder network for each task with a randomly initialized decoder. In this way, parameters of the ResNet-50 encoder are adjusted according to each task's characteristics. The encoder is essentially a task feature extractor, and we simply compute an FIM for this feature extractor. The FIM is then used as the task embedding for each task, which is a fixed dimensional vector.




\begin{algorithm}[t!]
\SetAlgoLined
    \caption{Calculate the approximation of MWAS}
    \label{alg: MWAS}
    \footnotesize
    \textbf{Input:}\\ 
    $A$: the adjacency matrix of a (cyclic) graph $\mathcal{G}$;\\
    $S$: the edge weight matrix;\\
    $\epsilon$: threshold, calculated by $\epsilon = 1-Acc(\tau_t)$;\\

    Set $s_0=a=0$, $r=b=||A||_1 $, $seg=b-a$;\\
    \While{$seg > 1$}{
    $A_T\leftarrow\textit{max-MAS}(A,r)$;\\ 
    \eIf{$A_T$ doesn't exist}{
    Find larger $r$: $r \leftarrow r + 1$;  \\
    Move the left endpoint of the interval to $r$: $a\leftarrow r$; \\
    }{         
    Calculate score: $s  \leftarrow \sum\limits_{i,j}(A_T \odot S)_{ij}$;\\
    \If{$R(A_T)<\epsilon$ and $s>s_0$}{
    Record maximal score: $s_0\leftarrow s$;\\
    Maintain a subgraph with maximal score: $A_{T}^{(best)}\leftarrow A_T$;  \\
    }
    Halve the length of the interval: $seg\leftarrow\lfloor \frac{seg}{2} \rfloor$; \\
    $r\leftarrow r-seg$;\\
    Move the right endpoint of the interval to $r$: $b\leftarrow r$; \\
    }
    }
    \textbf{Output:} $A_T^{(best)}$ as the approximation of the MWAS

\end{algorithm}

\subsection{Architecture relation graph ordering} \label{arch-graph prediction}
\noindent\textbf{Relation Graph Construction}
After obtaining the finetuned pairwise relation predictor on the target task $\tau_k$, we can construct a directed graph $\mathcal{G}^{\tau_k}$ with an adjacency matrix $A^{\tau_k}$. The presence of a directed edge from node $a$ to node $b$ in $\mathcal{G}^{\tau_k}$ represents the prediction that architecture $\mathit{arch}_a$ is better than architecture $\mathit{arch}_b$. That is, $A^{\tau_k}_{ij}=1$ indicates that $\mathit{arch}_a$ has higher performance than $\mathit{arch}_b$ in task $\tau_k$.
Since the predictor can be error-prone on the pairwise relation, there can be lots of noisy edges in the graph.
This can result in cycles that violate the transitivity of a partial order, which can affect the ranking of the models. (as in \cref{fig:overview}, $A_3 \to A_2 \to A_5 \to A_3$ forms a cycle). 
Ideally, we want to obtain a Directed Acyclic Graph (DAG), where its edges collectively define a partial order and its topological sort defines a ranking of nodes.

\noindent\textbf{Maximal Weighted Acyclic Subgraph (MWAS).} Based on our observations above, we aim to find a subgraph that satisfies the following properties: a) The edges exist with high confidence; b) There are no cycles in the subgraph; c) The subgraph is as close to the original graph as possible. This leads to our definition of Maximal Weighted Acyclic Subgraph (MWAS):
\vspace{-2mm}
\begin{definition}
(Maximal Weighted Acyclic Subgraph, MWAS) Given a directed (cyclic) graph $\mathcal{G}=(\mathcal{V},\mathcal{E})$, $|\mathcal{V}|=n$, with adjacency matrix $A$ and non-negative edge weight matrix $S$, the MWAS is the problem of finding an acyclic subgraph $T=(\mathcal{V},\mathcal{E_T})$ of $\mathcal{G}$ with adjacency matrix $A_T$, that maximizes the score $p(T)= \sum\limits_{e \in \mathcal{E}_T}S_e  = \sum\limits_{i,j}^n A_T \odot S $ and minimizes $||A_T-A||$.
\end{definition}
Intuitively, maximizing $p(T)$ enforces that if an edge exists, the network has high confidence in its correctness. Minimizing $||A_T-A||$ pushes the subgraph to keep as much edges from $\mathcal{G}$ as possible.
As Guo \etal mentioned \cite{DBLP:conf/icml/GuoPSW17}, the raw confidence values from a classifier can be poorly calibrated and are not reliable to determine the confidence of the classifier itself. Therefore, we adopt trust score defined in \cite{DBLP:conf/nips/JiangKGG18} to assign weights to the edges, determining to what extend we should trust an edge in $\mathcal{G}^{\tau_k}$.
\begin{definition}
(Trust Score) Given a testing sample $x \in \mathcal{X}$ and a trained classifier $h:\mathcal{X} \to \mathcal{Y}$, let $y_{pred}$ be the predicted class of $x$ and $y_n$ be the nearest class different from 
$y_{pred}$, then the trust score is defined by $\frac{d(x-y_n)}{d(x-y_{pred})}$.$^\textsuperscript{\ref{more details}}$
\end{definition}
With this definition, we calculate the trust score for each edge in Graph $\mathcal{G}$ and get the edge weight matrix $S$. 

The Maximal Acyclic Subgraph (MAS) problem was included by R.Karp in his list of 21 NP-complete problems.\cite{DBLP:conf/coco/Karp72}. Cvetkovic \etal proposed an algorithmic solution\cite{DBLP:journals/siammax/CvetkovicP20} to the max-MAS problem (\cref{def:maxMAS}) closely related to the MAS problem.
\begin{definition}
\label{def:maxMAS}
(The max-MAS problem) Finding the minimal integer $r$ such that a given graph $\mathcal{G}$ with adjacency matrix $A$ can be made acyclic by cutting at most $r$ incoming edges from each vertex.
\vspace{-2mm}
\end{definition}
Cvetkovic \etal formulated it as the following optimization problem (see S.M.\footnote{S.M. for Supplementary Materials\label{footnote S.M.}.}):
\begin{equation}
 \begin{aligned}
  & \min\,\, \rho(X) \\
  & s.t.\quad X \in \mathcal{B}(A,r)
 \end{aligned}
\end{equation}
where $\rho$ is the spectral radius and $\mathcal{B}(A,r)$ is the $L_1$ ball centered at $A$. Consequently, they find by integer bisection the smallest $r$ such that the objective function value equals to zero (a directed graph is acyclic if and only if its spectral radius is zero, see S.M.\textsuperscript{\ref{footnote S.M.}}). We denote the solution with respect to $r$ as \emph{max-MAS$(A,r)$}. We notice that in our relation graph where a large number of edges are clean, \emph{max-MAS$(A,r)$} can return reasonable solution $A_T$ in the measure of $||A_T-A||$ even when $r$ is not small enough. Suppose the accuracy on the validation set of target task $\tau_t$ is $Acc(\tau_t)$, we empirically save the \emph{max-MAS$(A,r)$} with edge dropping ratio $R(A_T) $ smaller than $1-Acc(\tau_t)$ during integer bisection of $r$ from $||A||_1$ to 0. With different approximations available, we calculate the trust score of these approximations and pick the one with the highest trust score as our approximation of MWAS. Details of the selection process are described in \cref{alg: MWAS}.


After obtaining a MWAS, we can apply transitive reduction to it and get a Hasse Digram. A topological sorting is easy to find out on a Hasse Digram and hence the predicted ranking of the models for this task is determined by the topological order.

\subsection{Training and Inference}
In a transfer learning setting, we first randomly sample a small budget of $m$ models from the source task and fully evaluate them. After the pairwise relation predictor is trained on the source task, we finetune it on each target task for another small budget of $b$ models. After finding the MWAS for each task's Arch-Graph, we evaluate the top $p$ models given by MWAS for each task and pick the best model as the final result. 
\section{Experiments}
\label{sec:experiments}
\begin{table*}[t]
    \small
    \centering
    \begin{tabular}{ll|ccccccc|c}
    \toprule
      \multicolumn{2}{c|}{Tasks}   &  Cls.O. & Cls.S. & Auto. & Normal & Sem. Seg. & Room.& Jigsaw & \multirow{2}{*}{Avg. Rank}\\
      \cmidrule{1-9}
      \multicolumn{2}{c|}{Metric}  & Acc.$^\uparrow$ & Acc.$^\uparrow$ & SSIM$^\uparrow$ & SSIM$^\uparrow$ & mIoU$^\uparrow$ & L2 loss$^\downarrow$ & Acc.$^\uparrow$ & \\
      \midrule
      \multirow{5}{*}{Single NAS}& RS\cite{DBLP:journals/jmlr/BergstraB12} & 46.85 & 56.50 & 70.06 & 60.70 & 28.37 & 59.35 & 96.78 & 59.26  \\

       &REA\cite{DBLP:conf/aaai/RealAHL19} & 47.09 & 56.57 & 69.98 & 60.88 & 28.87 & 58.73 & 96.88 & 41.03 \\
       &BONAS\cite{DBLP:conf/nips/ShiPXLKZ20} & 46.85 & 56.47 & 74.45 & 61.62 & 28.82 & 59.39 & 96.76 & 33.37 \\
       &weakNAS\cite{wu2021weak} & 47.40 & 56.88 & 72.54 & 62.37 & 29.18 & 57.86 & 96.86 & 10.49 \\
       & Arch-Graph-single &47.35 & 56.77 & 71.32 & 62.78 & 29.09 & 58.05 & 96.70 & 12.68   \\ 
       \midrule
      \multirow{8}{*}{Transfer NAS}& DT & 45.48 & 54.96 & 59.35 & 58.60 & 26.21 & 62.07 & 95.37 & 534.31 \\
     & CATCH\cite{DBLP:conf/eccv/ChenDCXCLZL20} & 47.29 & 56.49 & 70.36 & 60.85 & 28.71 & 59.37 & - & 37.72 \\

      & REA-t\cite{DBLP:conf/aaai/RealAHL19} & 46.98 & 56.60 & 73.41 & 61.02 & 28.90 & 58.18 & - & 28.98 \\
      & BONAS-t\cite{DBLP:conf/nips/ShiPXLKZ20} & 47.06 & 56.86 & 71.41 & 61.44 & 28.76 & 58.35 & - & 27.87 \\
      & nsganetv2\cite{DBLP:conf/eccv/LuDGBB20} & 46.86 & 56.29 & 73.77 & 61.41 & 28.73 & 59.07 & - & 34.39 \\
     & weakNAS-t\cite{wu2021weak} & 47.13 & 56.83 & 73.59 & 61.86 & 29.07 & 58.55 & - & 15.43 \\
     & Arch-Graph-zero &47.42 &56.78 &75.51 &63.39 &29.17 &58.15 & - &7.83\\
       &Arch-Graph & \textbf{47.44} & \textbf{56.98} & \textbf{75.90} & \textbf{64.35} & \textbf{29.19} & \textbf{57.75} & - & \textbf{5.24} \\
       \midrule
       & Global Best & 47.96 & 57.48 & 76.88 & 64.35 & 29.66 & 56.28 & 97.02 & 1\\
      \bottomrule
      
    \end{tabular}\\
    \footnotesize{ $^\uparrow$ indicates higher is better, $^\downarrow$ indicates lower is better, \textbf{bold} indicates the best result.}\vspace{-3mm}
    \caption{Performance comparisons between different NAS methods on our Arch-Graph on Macro level search space. Jigsaw results are omitted for TransferNAS methods because it is used as the pretrain task.}
    
    \label{tab:exp_tb101_macro}
    \vspace{-1mm}
\end{table*}

\subsection{Datasets and Implementation Details}
\noindent\textbf{TransNAS-Bench-101.}
TransNAS-Bench-101 (TB101) \cite{DBLP:conf/cvpr/DuanCXCLZL21} is a benchmark dataset providing architecture performance across seven vision domains including classification, regression, pixel-level prediction and self-supervised tasks. It provides opportunities to evaluate transferable NAS methods among different tasks.\footnote{More details can be found in Supplementary Materials\label{more details}}
There are two types of search space in this benchmark, i.e., the widely-studied cell-based search space containing 4096 architectures and macro skeleton search space based on residual blocks containing 3256 architectures.\textsuperscript{\ref{footnote S.M.}}
Following Lucaz \etal ~\cite{NEURIPS2020_768e7802}, we change the operation-on-edge setting in TransNAS-Bench-101 to an operation-on-node setting and encode each architecture as a graph with a fixed adjacency matrix and node feature matrix representing different operations.

\noindent\textbf{NAS-Bench-201.}
NAS-Bench-201 (NB201) \cite{DBLP:conf/iclr/Dong020} is a benchmark containing 15,625 architectures. It provides full information of these architectures on three classification tasks including CIFAR-10, CIFAR-100 and ImageNet-16-120.
Note that our Arch-Graph can also be applied to single-task setting. To further verify the effectiveness of our Arch-Graph, we conduct experiments of a single-task variant named \emph{Arch-Graph-single} by simply pretraining and finetuning the predictor on the same task.

\noindent\textbf{Pairwise Relation Predictor.}
To match the experiment in \cite{DBLP:conf/cvpr/DuanCXCLZL21}, we pretrain the pairwise relation predictor on the least time-consuming task, jigsaw (details of pretraining on other tasks can be found in S.M.\textsuperscript{\ref{footnote S.M.}}), restricting to a fixed budget of $m=50$ models. Then we finetune on each remaining task for another $b=30$ models using $b_f=20$ for training and $b_v=10$ for validation. Consequently, we construct the Arch-Graph using the predicted directed edges for each task and use them to get an ordering of architectures.

\noindent\textbf{Architecture Relation Graph Ordering.}
After we obtain architecture relation graph on the target tasks, we first use a naive method to order the architectures on the relation graph, named \textbf{Arch-Graph-zero}\footnote{More comparisons with comparator-based sorting algorithms can be seen in Sec. 5 in Supplementary Materials}. We implement the insertion sort algorithm to the model space by using the finetuned pairwise relation predictor as the comparison operator. Since there are incomparable elements and noisy edges (cycles) confusing the comparison operator, we simply skip the comparison until we can find a place to insert the not-yet-sorted architecture. This gives us a coarse ranking of the model space.

Because of high complexity of obtaining MWAS, we do not compute MWAS for the whole Arch-Graph. Instead, we pick top 500 models given by the coarse prediction of Arch-Graph-zero and construct the relation graph of 500 nodes using their predicted edges. Later ordering is conducted on this graph. After finding the MWAS (\cref{alg: MWAS}), we evaluate the top $p=20$ models given by the topological sort of these nodes. If any model selected for the final evaluation is already sampled, we simply skip it and evaluate the next model until we have evaluated $p$ models. Results on macro level search space and micro level search space can be found in \cref{tab:exp_tb101_macro} and \cref{tab:exp_tb101_micro}. 
\begin{table*}[t]
\small
    \centering
    \begin{tabular}{ll|ccccccc|c}
    \toprule
      \multicolumn{2}{c|}{Tasks}   &  Cls.O. & Cls.S. & Auto. & Normal & Sem. Seg. & Room.& Jigsaw & \multirow{2}{*}{Avg. Rank}\\
      \cmidrule{1-9}
      \multicolumn{2}{c|}{Metric}  & Acc.$^\uparrow$ & Acc.$^\uparrow$ & SSIM$^\uparrow$ & SSIM$^\uparrow$ & mIoU$^\uparrow$ & L2 loss$^\downarrow$ & Acc.$^\uparrow$ & \\
      \midrule
      \multirow{5}{*}{Single NAS}& RS\cite{DBLP:journals/jmlr/BergstraB12} & 45.16 & 54.41 & 55.94 & 56.85 & 25.21 & 61.48 & 94.47 & 85.61  \\
      
       & REA\cite{DBLP:conf/aaai/RealAHL19} & 45.39 & 54.62 & \textbf{56.96} & 57.22 & 25.52 & 61.75 & 94.62 & 38.50 \\
       & BONAS\cite{DBLP:conf/nips/ShiPXLKZ20} & 45.50 & 54.46 & 56.73 & 57.46 & 25.32 & 61.10 & 94.81 & 34.31 \\ 
       & weakNAS\cite{wu2021weak} & 45.66 & 54.72 & 56.77 & 57.21 & \textbf{25.90} & 60.31 & 94.63 & 20.03 \\ 
       & Arch-Graph-single & 45.48 & 54.70 & 56.52 & 57.53 & 25.71 & 61.05 & 94.66 & 22.15 \\ 
       \midrule
      \multirow{8}{*}{Transfer NAS}& DT & 42.03 & 49.80 & 51.20 & 55.03 & 22.45 & 66.98 & 88.95 & 935.12 \\
     & CATCH\cite{DBLP:conf/eccv/ChenDCXCLZL20} & 45.27 & 54.38 & 56.13 & 56.99 & 25.38 & 60.70 & - & 63.49 \\
      & REA-t\cite{DBLP:conf/aaai/RealAHL19} & 45.51 & 54.61 & 56.52 & 57.20 & 25.46 & 61.04 & - & 40.14 \\
      & BONAS-t\cite{DBLP:conf/nips/ShiPXLKZ20} & 45.38 & 54.57 & 56.18 & 57.24 & 25.24 & 60.93 & - & 55.30 \\
      & nsganetv2\cite{DBLP:conf/eccv/LuDGBB20} & 45.61 & 54.75 & 56.47 & 57.24 & 25.36 & 61.73 & - & 34.89 \\
     & weakNAS-t\cite{wu2021weak} & 45.29 & 54.78 & 56.90 & 57.19 & 25.41 & 60.70 & - & 35.73 \\
     & Arch-Graph-zero & 45.64 & 54.80 & 56.61 & 57.90 & 25.73 & 60.21& & 14.7 \\
       &Arch-Graph & \textbf{45.81} &\textbf{54.90} & 56.58 & \textbf{58.27} & 25.69 & \textbf{60.08} & -&   \textbf{12.2}\\
       \midrule
       & Global Best & 46.32 & 54.94 & 57.72 & 59.62 & 26.27 & 59.38 & 95.37 & 1\\
      \bottomrule
      
    \end{tabular}\\
    \footnotesize{ $^\uparrow$ indicates higher is better, $^\downarrow$ indicates lower is better, \textbf{bold} indicates the best result.}\vspace{-3mm}
    \caption{Performance comparisons between different NAS methods and our Arch-Graph on Micro level search space. Jigsaw results are omitted for TransferNAS methods because it is used as the pretrain task.}
    
    \label{tab:exp_tb101_micro}
    \vspace{-1mm}
\end{table*}

\subsection{Comparison with state-of-the-art NAS}

\noindent\textbf{Single-task NAS.}
On TB101, we use Random Search (RS)\cite{DBLP:journals/jmlr/BergstraB12} and Regularized Evolutionary Algorithm (REA)\cite{DBLP:conf/aaai/RealAHL19} for 50 epochs as baselines.
We then conduct experiments using two state-of-the-art predictor-based NAS methods, BONAS\cite{DBLP:conf/nips/ShiPXLKZ20} and weakNAS\cite{wu2021weak} on each task. The total budget for each method is set to 50 randomly selected models. The average model rank is averaged across six target tasks. As in \cref{tab:exp_tb101_macro,tab:exp_tb101_micro}, weakNAS is the best in single-task setting and Arch-Graph-single achieves comparable results to weakNAS. On NB201, we conduct experiments on CIFAR-100 (\cref{correlation coefficients and budgets}) and set the budgets to 150 models. Better than REA and RS, Arch-Graph has an average performance of 73.38\% that outperforms BONAS. Although slighter lower than weakNAS, Arch-Graph has a much larger kendall-rank coefficient (0.67) than weakNAS (0.49), indicating a better ordering of the whole model space. 

\noindent\textbf{Task-Transferrable NAS.}
The transferred version of weakNAS and BONAS are also pretrained on jigsaw with a budget of 50 models. After initializing the predictors, we sample another 50 models to finetune the GCN embedding extractor and Bayesian Sigmoid Regression in BONAS and sets of the weak predictors in weakNAS on the target task. In addition to the searched models' accuracy, we also report the model rank in the search space, averaged across 6 targeted tasks (\cref{correlation coefficients and budgets}). Our Arch-Graph shows great superiority over both single task methods and transferable NAS methods when transferring knowledge from a pre-trained predictors, surpassing weakNAS\cite{wu2021weak} by average model rank 10.19 on macro level search space and 23.53 on micro level search space. It takes at least 60\% extra samples for other methods to achieve comparable results, in \cref{correlation coefficients and budgets}.

\begin{table}[t]
\small
    \setlength{\tabcolsep}{4pt}

    \centering
    \begin{tabular}{ll|cccc}
    \toprule
           & Methods& $\tau^\uparrow$ & $\rho^\uparrow$   & \#budgets$^\downarrow$\\
           \midrule
          \multirow{7}{*}{TB101} &BONAS\cite{DBLP:conf/nips/ShiPXLKZ20}  & 0.26  & 0.38  & 100+\\
           & BONAS-t\cite{DBLP:conf/nips/ShiPXLKZ20} &0.24  &0.34 & 100+\\
           &nasganetv2\cite{DBLP:conf/eccv/LuDGBB20} &0.19 &0.28 & 100+ \\
           &weakNAS\cite{wu2021weak} &0.36 &0.51 &  80\\
         &weakNAS-t\cite{wu2021weak} &0.16 &0.24 &  100\\
         &Arch-Graph-zero &0.58 &0.76 & 60\\
         &Arch-Graph& \textbf{0.61} & \textbf{0.79} &  \textbf{50}\\
         \midrule
          &Methods & Acc.$^\uparrow$ & $\tau^\uparrow$ & $\rho^\uparrow$ \\
         \midrule
         \multirow{5}{*}{NB201} &RS\cite{DBLP:journals/jmlr/BergstraB12}& 71.80 & -  & - & \\
         &REA\cite{DBLP:conf/aaai/RealAHL19} &72.70 &- & - & \\
         &BONAS\cite{DBLP:conf/nips/ShiPXLKZ20} &72.84& 0.43& 0.60 \\
         &weakNAS\cite{wu2021weak} & \textbf{73.42}&0.49& 0.56 &\\
         &Arch-Graph& 73.38 & \textbf{0.67} & \textbf{0.79}  & \\
    \bottomrule
    \end{tabular}\vspace{-3mm}
    \caption{Comparison of different methods on TransNAS-Bench-101 and NAS-Bench-201 benchmarks. $\tau,\rho$ are Kendall rank coefficient, Pearson correlation coefficient respectively. \#budgets indicates the number of architectures for a method to find top 0.3\% architectures in the macro level search space.}
    \vspace{-3mm}
    \label{correlation coefficients and budgets}
    \vspace{-4mm}
\end{table}
\begin{figure*}
    \centering
    \vspace{-1em}
    \includegraphics[width=2.0\columnwidth]{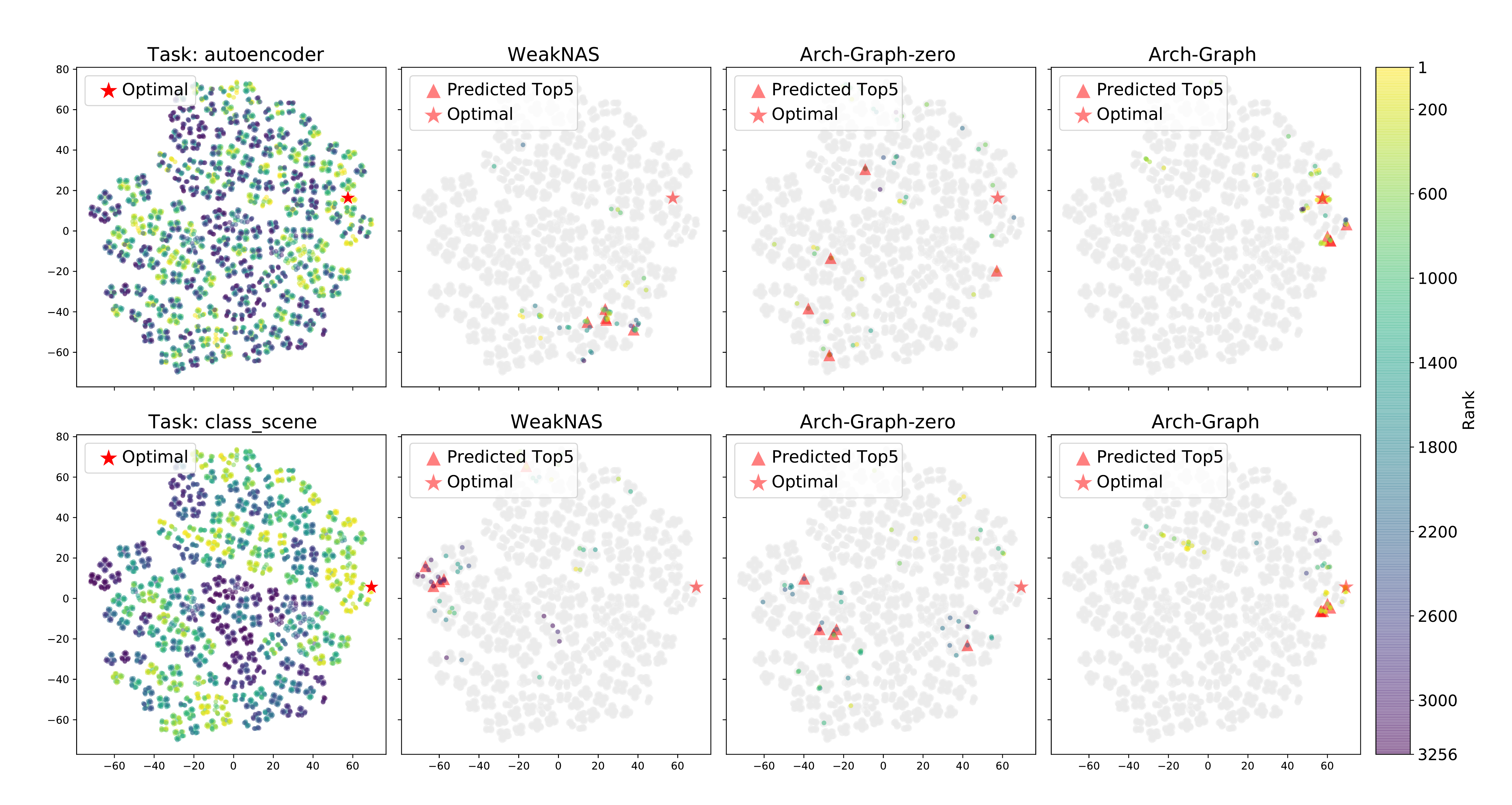}\vspace{-5mm}
    \caption{Visualization of the network search space on object classification and autoencoding tasks. For each algorithm, we color its predicted top-50 models and grey out everything else. We use triangles to mark each algorithm's top-5 prediction, and use stars to label the search space's global optima.}\vspace{-3mm}
    \label{fig:visualization}
\end{figure*}

To better illustrate the effectiveness of our Arch-Graph, in \cref{fig:visualization}, we show the visualization result of the predicted top 50 models in the macro level search space on two tasks. More visualizations of search results on other tasks can be found in S.M.\textsuperscript{\ref{footnote S.M.}}. We first use t-SNE to project the model into the 2-dimensional space and colors to indicate model performance. The shallower the color, the stronger the model. In this projection, top models for each task tend to form local clusters. WeakNAS and Arch-Graph-zero can both attend to local optima, whereas Arch-Graph's predictions are significantly closer to globally optimal architectures.


\subsection{Ablation Study}
\label{ablation}
\noindent\textbf{Task embedding.} Some works on transferable NAS\cite{DBLP:conf/nips/WongHLG18} also propose to use task embeddings to guide the search when facing different tasks. However, they use a randomly initialized embedding to represent each task and it is learned jointly with the NAS model's parameters. We verify the effectiveness of our task embedding defined in \cref{sec:methods}. We compare our task embedding with randomly initialized vectors for each task's embedding. We show the averaged architecture rank over 6 target tasks, with experiments repeated over 5 random seeds in \cref{ablation_task2vec}. The performance using randomly initialized task embedding is highly unstable, resulting in a much larger variance (0.63 vs 692.03) and a significantly lower average performance (24.13) compared to Task2Vec (5.24), indicating a randomly initialized task embedding can't guarantee a stable knowledge transfer. 

\noindent\textbf{MWAS.}
Obtaining the approximation of the Maximal Weighted Acyclic Subgraph Problem is a central component of our model to improve graph construction.
To show its advantages over Arch-Graph-zero, we first pick 20 finetuned predictors on each task with the highest validation accuracy among the $b_v$ validation architectures. We then compare the predicted accuracy between Arch-Graph-zero and Arch-Graph. Arch-Graph can identify better models than Arch-Graph-zero, which on average improves the rank by 3.14 and 5.28 on the macro and micro search space, respectively. More detailed differences of these top models can be found in \cref{tab:exp_tb101_macro,tab:exp_tb101_micro}.

\noindent\textbf{Arch-Graph-single.}
To verify the effect of knowledge transfer from source tasks to new target tasks, we compare the performance of Arch-Graph and Arch-Graph-single and fix the total budget to 50 models. Compared to transferring knowledge from a pretrained predictor, Arch-Graph-single is worse than Arch-Graph as shown in \cref{tab:exp_tb101_macro,tab:exp_tb101_micro}. It shows the effectiveness of knowledge transfer from predictor trained on a previous task.

\begin{table}[t!]
\small
    \setlength{\tabcolsep}{12pt}

    \centering
    \begin{tabular}{l|cc}
    \toprule
           Average rank& Mean & Variance \\
           \midrule
           Ours& 5.24 & 0.63\\
         Random  & 24.13 & 692.03\\
    \bottomrule
    \end{tabular}\vspace{-3mm}
    \caption{Searched network's rank comparison by two embedding methods on Arch-Graph (lower is better).}
    \vspace{-5mm}
    \label{ablation_task2vec}
\end{table}

\section{Conclusions and Discussions}
In this work, we propose Arch-Graph, a task-transferable NAS method that formulate NAS as a graph ordering problem on an architecture relation graph. Directed edges of this graph are obtained through training a pairwise relation predictor with knowledge transfer. With extensive experiment, we demonstrate Arch-Graph's transferability and sample efficiency over many other NAS methods.


\noindent\textbf{Potential negative societal impact.} We have not identified any potential negative social impact. All the datasets we use are public and conform with ethical standards.

\noindent\textbf{Limitation and Future Work.}
With Arch-Graph-zero, it is possible to exclude the ground truth global optima before the MWAS calculation. Future work could explore along this direction and construct subgraphs more efficiently for ranking. For example, the pairwise relation predictor training and the MWAS calculation can be done in an iterative style, so that we can progressively shrink the search space and improve the performance.

\section{Acknowledgements}
\fontdimen2\font=0.2ex
This work was supported in part by National key R\&D Program of China under Grant No.2020AAA0109700, National Natural Science Foundation of China (NSFC) No.61976233, Guangdong Province Basic and Applied Basic Research (Regional Joint Fund-Key) Grant No.2019B1515120039, Guangdong Outstanding Youth Fund (Grant No. 2021B1515020061), Shenzhen Fundamental Research Program (Project No.RCYX20200714114642083, No.JCYJ20190807154211365).

{\small
\bibliographystyle{ieee_fullname}
\bibliography{egbib}
}

\end{document}